\documentclass[runningheads]{llncs}
\usepackage{amssymb}

\usepackage{graphics}
\usepackage{epsfig}

\usepackage{multirow}

\usepackage{subfigure}
\usepackage{graphicx}
\usepackage{amsmath}
\usepackage{color}
\usepackage{arydshln}
\usepackage{rotating}
\begin{document}
\title{Holistic Decomposition Convolution for Effective Semantic Segmentation of 3D MR Images}
%
%
\author{Guodong Zeng\inst{1}
 \index{Zeng, Guodong}
\and
Guoyan Zheng\inst{1,2}
	\index{Zheng, Guoyan}
}

\institute{ 
  Institute of Surgical Technology \& Biomechanics, University of Bern, Switzerland. \ \\
	\texttt{Correspondence: guoyan.zheng@ieee.org}
	\and
	School of Biomedical Engineering, Shanghai Jiao Tong University, No.800 Dongchuan Road, 200240 Shanghai, China
}
\maketitle              
\begin{abstract}
Convolutional Neural Networks (CNNs) have achieved state-of-the-art performance in many different 2D medical image analysis tasks. In clinical practice, however, a large part of the medical imaging data available is in 3D. This has motivated the development of 3D CNNs for volumetric image segmentation in order to benefit from more spatial context. Due to GPU memory restrictions caused by moving to fully 3D, state-of-the-art methods depend on subvolume/patch processing and the size of the input patch is usually small, limiting the incorporation of larger context information for a better performance. In this paper, we propose a novel Holistic Decomposition Convolution (HDC), for an effective and efficient semantic segmentation of volumetric images. HDC consists of a periodic down-shuffling operation followed by a conventional 3D convolution. HDC has the advantage of significantly reducing the size of the data for sub-sequential processing while using all the information available in the input irrespective of the down-shuffling factors. Results obtained from comprehensive experiments conducted on hip T1 MR images and intervertebral disc T2 MR images demonstrate the efficacy of the present approach.

\end{abstract}
\section{Introduction}
Automated segmentation of volumetric medical images is a challenging task. The more recent development of deep neural networks, and in particular convolutional neural networks (CNN) \cite{NIPS2012_4824}, suggests a course of methods \cite{Litjens_MedIA_2017}. Contrary to conventional shallow learning methods, where feature design is crucial, deep learning methods automatically learn hierarchies of relevant features directly from the training data. Early works \cite{Prasson_MICCAI_2013} treat the image segmentation as a classification problem with sliding window where CNNs are applied to input patches to classify the central pixel/voxel of each patch. As classifying each pixel/voxel in a sliding window fashion results in orders of magnitude of redundant calculation, most of recent works \cite{Ronneberger_MICCAI_2015,Kamnitsas_MedIA_2017} are based on Fully Convolutional Networks (FCNs) \cite{Long_CVPR_2015}, which can process the input image in an end-to-end way and can provide a full resolution segmentation map \cite{Litjens_MedIA_2017}. In several biomedical image segmentation benchmarking competitions, methods built on CNNs \cite{Ronneberger_MICCAI_2015,Zheng_MedIA_2017} are on the top list of the associated leaderboard. Despite the fact that CNNs-based methods have achieved state-of-the-art performance in many different 2D medical image analysis tasks, in clinical practice, however, a large part of the medical imaging data available is in 3D. This has motivated the development of 3D CNNs for volumetric image segmentation in order to benefit from more spatial context. For example, Kamnitsas et al. \cite{Kamnitsas_MedIA_2017} proposed a dual pathway, 11 layers deep 3D multi-scale CNN with fully connected Conditional Random Field (CRF) for brain lesion segmentation and achieved state-of-the-art performance. {\c{C}}i{\c{c}}ek et al. \cite{3DUNET_2016} proposed the 3D U-net as an extension to the 2D U-net by replacing all 2D operations with their 3D counterparts. By incorporation of residual blocks and using a similar architecture as the 3D U-net, Milletari et al. \cite{Milletari_VNet_2016} proposed the 3D V-net for volumetric medical image segmentation. One thing common to all these 3D CNNs-based approaches is that they all follow a fully convolutional downsample-upsample pathway. More specifically, the downsampling path tries to achieve higher-level feature abstraction by gradually downsampling low-level features with high spatial resolutions while the upsampling path aims to upsample the learned high-level features to achieve a full-resolution segmentation. Deviating from the fully convolutional downsample-upsample pathway, Li et al. \cite{Li_IPMI_2017} proposed a high-resolution network architecture referred as ``HighRes3DNet'' for the segmentation of fine structure in volumetric images. HighRes3DNet preserves the spatial resolution throughout the layers and the enlargement of the receptive field is then achieved by incorporating dilated convolution.

Due to GPU memory restrictions caused by moving to fully 3D, state-of-the-art methods \cite{3DUNET_2016,Milletari_VNet_2016,Kamnitsas_MedIA_2017,Li_IPMI_2017} depend on subvolume/patch processing. The size of the input patch is usually small if no specialized hardware with large GPU memory is used, limiting the incorporation of larger context information for a better performance. To tackle these challenges, we present a novel and efficient approach which allows for using large size of patches for an effective and efficient semantic segmentation of volumetric images by leveraging context information in a large patch. Our contributions can be summarized as follows:

\begin{itemize}
	\item First, we propose a novel Holistic Decomposition Convolution (HDC), which can be regarded as an inverse operation to the previously introduced Dense Upsampling Convolution (DUC) \cite{Shi_CVPR_2016,Wang_WACV_2018}. HDC consists of a periodic down-shuffling operation followed by a conventional 3D convolution. HDC has the advantage of significantly reducing the size of the data for sub-sequential processing while using all the information available in the input irrespective of the down-shuffling factors. We apply HDC directly to the input data, whose output will be used as the input for sub-sequential CNNs. In order to achieve volumetric dense prediction at final output, we need to recover full resolution, which is done by using DUC. 
	\item Second, we extensively validate the proposed approach on the task of segmentation of hip bony structures from T1 MR images of limited field of view. We show that HDC and DUC are network agnostic and can be combined with different FCNs for an improved performance. More specifically, we demonstrate that the improved performance can be obtained when HDC and DUC are used with 3D U-net\cite{3DUNET_2016}, 3D V-net \cite{Milletari_VNet_2016}, and HighRes3DNet \cite{Li_IPMI_2017}, respectively. We investigate the influence of the down-shuffling factors on the segmentation results.
	\item Third, in addition to the hip MR image segmentation task, we apply the proposed approach off-the-shelf to a typical yet highly challenging segmentation task, i.e., intervertebral disc (IVD) segmentation from T2 MR images. We conduct comprehensive cross-validation experiments on an open dataset to compare the performance of our approach with that of state-of-the-art methods. We have achieved better segmentation results than state-of-the-art methods.
\end{itemize}

\section{Methods}
In this section, we will first briefly present the usage of DUC for semantic segmentation, followed by a detailed description of HDC. We will then show how to combine HDC and DUC with FCNs for effective segmentation of volumetric images. 

\subsection{Dense Upsampling Convolution for Semantic Segmentation}
For a typical FCNs-based approach that follows the downsample-upsample pathway, in order to achieve volumetric dense prediction, we need to recover full resolution at output. Conventional methods such as bilinear upsampling \cite{Yu_CVPR_2017} is not attractive as the upsampling parameters are not learnable. Deconvolution could be an alternative but, unfortunately, it can easily lead to ``uneven overlap'', resulting in checkboard artifacts \cite{Aitken_ARXIV_2017}. In \cite{Shi_CVPR_2016}, DUC, which consists of low-resolution convolution with a periodic up-shuffling operator, was proposed to jointly learn the feature extraction and upsampling weights for super-resolution reconstruction. DUC was later used as the last layer for semantic segmentation in \cite{Wang_WACV_2018}. For details about DUC, we refer to previouw work \cite{Shi_CVPR_2016,Wang_WACV_2018}.

\begin{figure}[tbp]
\centering
\includegraphics[width=\textwidth]{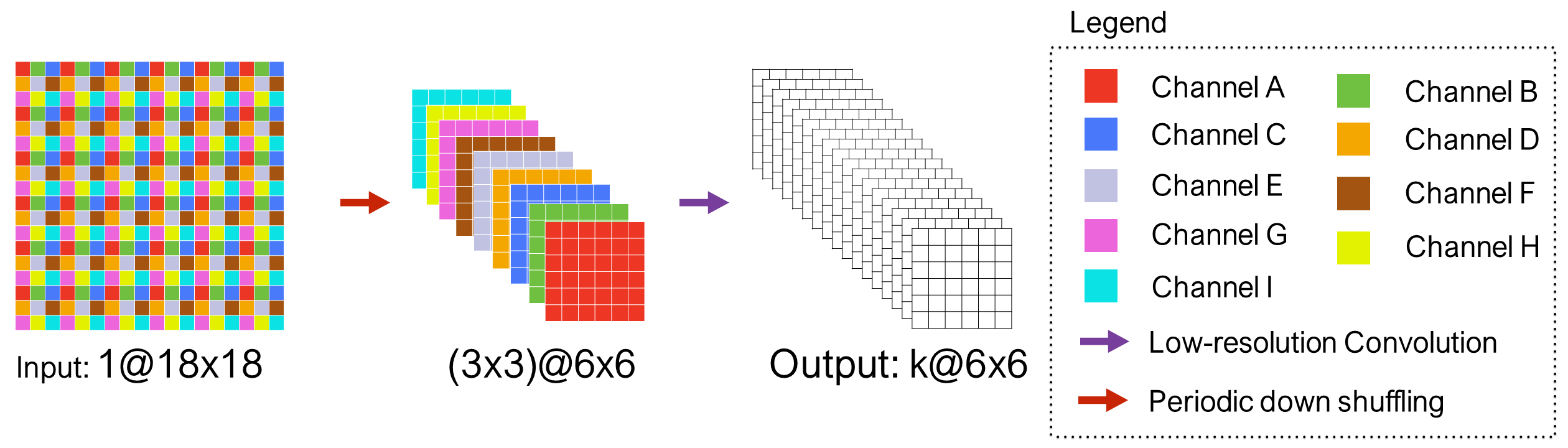}
\caption{A schematic view of holistic decomposition convolution in 2D which consists of a periodic down-shuffling operation with low-resolution convolution. Here the down-shuffling factor is (3, 3). }
\label{fig:HDC}
\end{figure} 

\subsection{Holistic Decomposition Convolution}
HDC can be regarded as the inverse operation to DUC and as shown in Fig. \ref{fig:HDC}, it consists of a periodic down-shuffling operator with low-resolution convolution. HDC is designed to be directly applicable to the input data with the aim to reduce the size of the data for sub-sequential processing while using all the information available in the input irrespective of the down-shuffling factors. This is also the reason why we call this novel operation as ``Holistic Decomposition Convolution''. More specifically, let's assume that the size of input data ($I^{HR}$) is $(\mathnormal{n_x} \times d) \times (\mathnormal{n_y} \times h) \times (\mathnormal{n_z} \times w) \times C$ and the size of the output from HDC is $d \times h \times w \times k$, where $\mathnormal{n_x}, \mathnormal{n_y}, \mathnormal{n_z}$ are the down-shuffling factors along the three spatial axes, respectively; $C$ is the number of channels in the input data; $k$ is the number of feature maps in the output of HDC. Instead of applying convolution to high resolution (HR) images, we first apply a periodic down-shuffling operator to the input data to get $C \times (\mathnormal{n_x} \times \mathnormal{n_y} \times \mathnormal{n_z})$ channels of feature maps with low resolution (LR) and then apply convolutions with a kernel size of $3 \times 3 \times 3$ to get the $k$ feature maps of size $(d \times h \times w)$. Mathematically, this can be described as:

\begin{equation}\label{eq:dwsc}
\begin{aligned}
HDC(I^{LR}; W_1, b_1) = \phi (W_1 * \mathnormal{PDS} (I^{HR}) + b_1) 
\end{aligned}
\end{equation}

where $\phi$ is an non-linear activation function that is applied element-wise; $W_1$, $b_1$ are trainable weights and bias, respectively; $\mathnormal{PDS}$ is a periodic down-shuffling operator which aims to rearrange the tensor ($T_{HR}$) in the shape of $(\mathnormal{n_x} \times d) \times (\mathnormal{n_y} \times h) \times (\mathnormal{n_z} \times w) \times C$ to the tensor ($T_{LR}$) in the shape of $(d \times h \times w) \times (C \times (\mathnormal{n_x} \times \mathnormal{n_y} \times \mathnormal{n_z}))$. And the operation $T_{LR} = \mathnormal{PDS}(T_{HR})$ can be mathematically described as below:

\begin{equation}\label{eq:dws}
\begin{aligned}
T_{LR}(x',y',z',c') = T_{HR}( {x' \cdot n_x}+\left \lfloor{mod(c', n_x\cdot C)/C}\right \rfloor, \\
{y' \cdot n_y}+\left \lfloor{mod(c',n_x n_y \cdot C)/(n_x\cdot C)}\right \rfloor, \\
{z' \cdot n_z}+\left \lfloor{c'/(n_x n_y \cdot C)}\right \rfloor, \\
mod(c', C) \ )
\end{aligned}
\end{equation}

where $x',y',z',c'$ are the coordinates of the voxels in the low resolution space, and $x' \in [0,d-1], y' \in [0,h-1], z' \in [0, w-1], c' \in [0, C \cdot \mathnormal{n_x} \cdot \mathnormal{n_y} \cdot \mathnormal{n_z}-1] $. 

\begin{figure}[tbp]
\centering
\includegraphics[width=\textwidth]{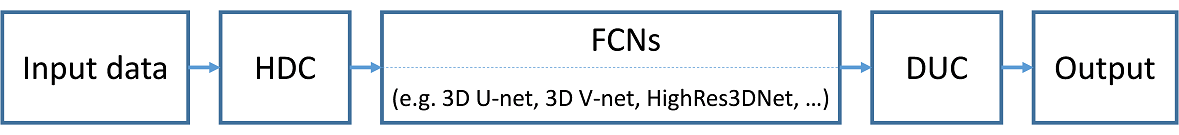}
\caption{A schematic view of how to augment existing FCNs with HDC and DUC for semantic segmentation. }
\label{fig:HDC_DUC_SemanticSeg}
\end{figure}

\begin{figure}[tbp]
\centering
\includegraphics[width=\textwidth]{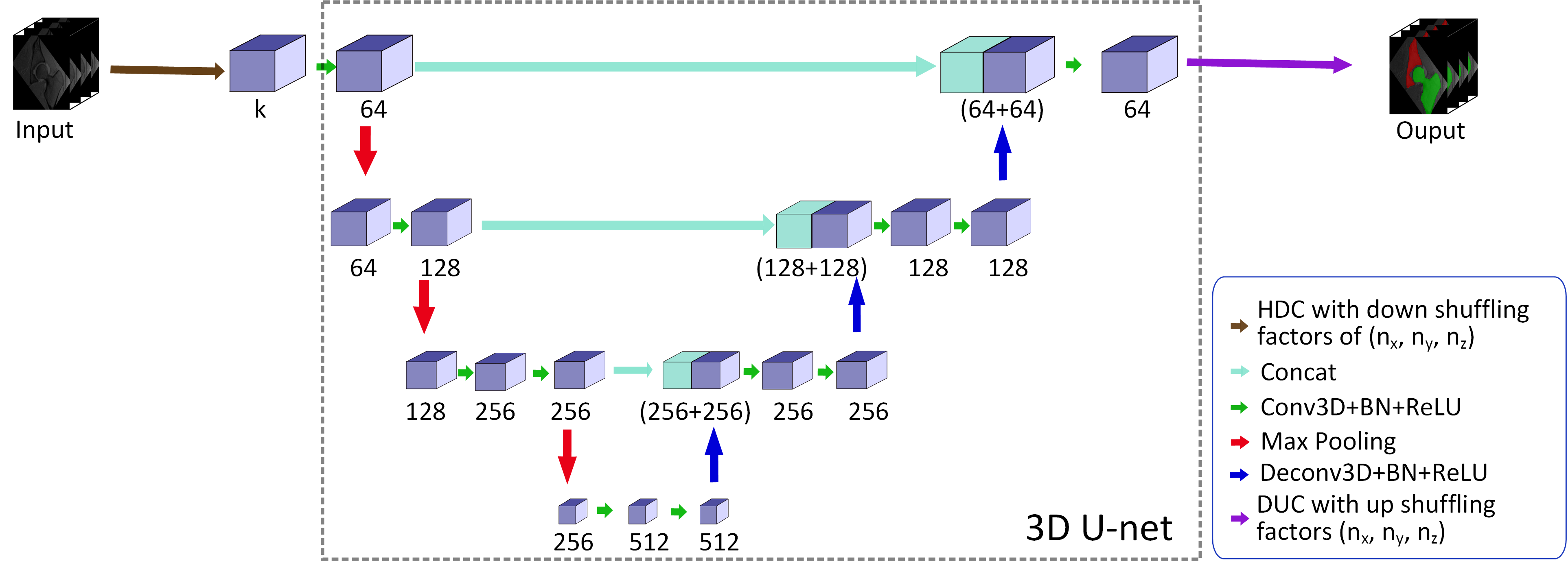}
\caption{A schematic view of how to augment 3D U-net with HDC and DUC for segmenting 3D hip MR images of limited field of view. The numbers below each block represent the number of feature maps.}
\label{fig:3D-LP-U-net}
\end{figure}

\subsection{HDC and DUC Augmented FCNs for Volumetric Image Segmentation}
Both DUC and HDC are network agnostic and can be combined with existing FCNs such as 3D U-net \cite{3DUNET_2016}, 3D V-net \cite{Milletari_VNet_2016}, and HighRes3dNet \cite{Li_IPMI_2017} for semantic segmentation as shown in Fig. \ref{fig:HDC_DUC_SemanticSeg}, as long as the dimensions of the output from HDC satisfy the input requirement of the deep neural networks. Fig. \ref{fig:3D-LP-U-net} shows an example of combining HDC and DUC with 3D U-net for segmenting 3D hip MR images of limited field of view. The advantage of such a pipeline is apparent. When a HDC with down-shuffling factors of $(\mathnormal{n_x}, \mathnormal{n_y},  \mathnormal{n_z})$ is applied to the input data, both the computational and the storage cost for the underlying 3D U-net will be reduced by a factor of $(\mathnormal{n_x} \cdot \mathnormal{n_y} \cdot \mathnormal{n_z})$, allowing one to use large patch as the input. The full resolution segmentation map is then obtained at the final output by applying a DUC with up-shuffling factors of $(\mathnormal{n_x}, \mathnormal{n_y}, \mathnormal{n_z})$. To differentiate from the original 3D U-net, we call the 3D U-net augmented with HDC and DUC as 3D large patch U-net (3D LP-U-net). Similarly we can derive 3D LP-V-net and LP-HighRes3DNet respectively by augmenting the original 3D V-net \cite{Milletari_VNet_2016} and HighRes3DNet \cite{Li_IPMI_2017} with HDC and DUC. In this study, we take the original 3D U-net, 3D V-net and HighRes3DNet as the baseline to evaluate the performance of the associated networks augmented with HDC and DUC. For all the studies, a combination of cross entropy loss with Dice loss as introduced in \cite{Milletari_VNet_2016} is used.

\subsection{Implementation Details}
All methods reported in this study were implemented in Python using TensorFlow framework and were trained and tested on a desktop with a 3.6 GHz Intel(R) i7 CPU and a NVIDIA GTX 1080 Ti graphics card with 11 GB GPU memory. We empirically fixed the number of output feature maps from the HDC as $k=64$. All the weights were initialized from Gaussian distribution ($\mu  = 0$, $\sigma  = 0.01$) and were then updated by the stochastic gradient descent (SGD) algorithm (momentum = 0.9, weight decay = 0.005). For the baseline networks, the initial learning rate was chosen to be 0.001 and was halved every 3000times iterations. For the networks augmented with HDC and DUC, depending on the shuffling factors, different initial learning rates were used as described below. During a training stage, we randomly cropped sub-volume patches of a fixed size from training samples. Each sampled patch was normalized as zero mean and unit variance before fed into network. During a testing stage, given a test volumetric image, we extracted overlapped sub-volume patches with the same size as we used in the associate training stage and fed them to the trained network to get prediction probability maps. For the overlapped voxels, the final probability maps would be the average of the overlapped patches, which were then used to derive the final segmentation results.

\section{Experiments and Results}
In this section, we present experimental results of the proposed pipeline for volumetric image segmentation. Two datasets, i.e., an in-house dataset consisting of 25 T1 hip MR images with limited field of view and a publicly available dataset from the MICCAI 2015 IVD localization and segmentation challenge \cite{Zheng_MedIA_2017}, are used in our study. More specifically, first, we conduct an ablation study on the in-house hip dataset to evaluate the influence of the shuffling factors and of the underlying FCNs on the performance of the proposed pipeline. Based on the findings from the ablation study, we choose the 3D LP-U-net for our remaining studies. Following \cite{Zheng_MedIA_2017}, we used Dice Overlap Coefficients (DOC), Average Surface Distance (ASD) and Hausdorff Distance (HD) as the evaluation metrics.

\subsection{Ablation study on hip MR images with limited field of view}
\subsubsection{Data and augmentation}
In this study, we used 25 3D T1 MR images, acquired from patients with hip pain. Those images were acquired by using a dual-flip angle 3D gradient-echo technique (TR/TE = 15/3.3 ms; flip angles: $4^o$ and $24^o$; slice thickness: 1.0mm; field of view: $160 \times 160$ $mm^2$). All images were resampled to have a uniform size of $480 \times 480 \times 160$ voxels with an average voxel spacing of 0.374mm $\times$ 0.363mm $\times$ 1.078mm. Slice by slice manual segmentation was used to create the reference ground truth segmentation. We randomly distributed the 25 datasets into two groups with one group containing 20 datasets as the training data and the remaining 5 datasets as the testing data. During training, data augmentation was used to enlarge the training samples. Specifically, we applied a smooth deformation field on both data and ground truth labels. For this, we sampled random vectors from a normal distribution with a standard deviation of 15 voxels in a $2 \times 2 \times 2$ grid of control points and then applied a B-spline interpolation. For each training sample, we generated four additional augmented samples. All the networks used in this study were trained on the augmented training data for 10,000 iterations.

\begin{table}[tbp]
\centering
\caption{Results of investigation of different patch sizes on the performance of the original 3D U-net. Ace: acetabulum; Femur: the proximal femur}
\label{Tab:3DU-net:PatchSizes}
\resizebox{\textwidth}{!}
{
\begin{tabular}{l|c|c|c|c|c|c} \hline
Patch size & \multicolumn{2}{c|}{(50, 50, 40)}        & \multicolumn{2}{c|}{(96, 96, 96)}     & \multicolumn{2}{c}{(200, 200, 40)}  \\ \hline
Anatomy    & Ace                  & Femur            & Ace              & Femur             & Ace              & Femur            \\ \hline
DOC (\%)   & 37.45 $\pm$ 5.73     & 30.62 $\pm$ 3.55 & 91.30 $\pm$ 5.84 & 95.89 $\pm$ 1.21  & 92.06 $\pm$ 5.37 & 96.84 $\pm$ 0.90 \\ \hline
ASD (mm)   & 28.15 $\pm$ 5.04     & 29.27 $\pm$ 4.90 & 5.11 $\pm$ 7.57  & 1.41 $\pm$ 1.11   & 0.88 $\pm$ 0.76  & 0.63 $\pm$ 0.31  \\ \hline
HD (mm)    & 111.10 $\pm$ 10.41 & 95.1 $\pm$ 8.98  & 33.92 $\pm$ 29.0 & 29.78 $\pm$ 20.35 & 13.71 $\pm$ 5.07 & 10.85 $\pm$ 6.09 \\ \hline
\end{tabular}
}
\end{table}

\subsubsection{Ablation study}
We first investigated the influence of patch sizes on the performance of the original 3D U-net. The results are presented in Table \ref{Tab:3DU-net:PatchSizes}. It was observed that better performance was obtained when larger patch size was used. Due to the GPU memory constraint, $200 \times 200 \times 40$ is the maximum size that we can use.

We then examined the effect of different shuffling factors on the performance of the 3D LP-U-net when a fixed patch size of $400 \times 400 \times 80$ was used. The results are reported in Table \ref{Tab_3DLPU-net_ShufflingFactors}. From this table, we can see that (1) the higher the shuffling factors, the bigger the initial learning rate that we used; (2) the higher the shuffling factor, in general the less accurate the results but the best results were achieved when the shuffling factor was (4, 4, 2); (3) even with a shuffling factor as high as (25, 25, 2), we still get sub-millimeter segmentation accuracy for both the acetabulum and the proximal femur; and (4) in comparison with the results reported in Table \ref{Tab:3DU-net:PatchSizes}, 3D LP-U-net achieved better results than the original 3D U-net with the largest patch size when the shuffling factor was smaller than (16, 16, 2).

\begin{figure}[tbp]
\centering
\includegraphics[width=\textwidth]{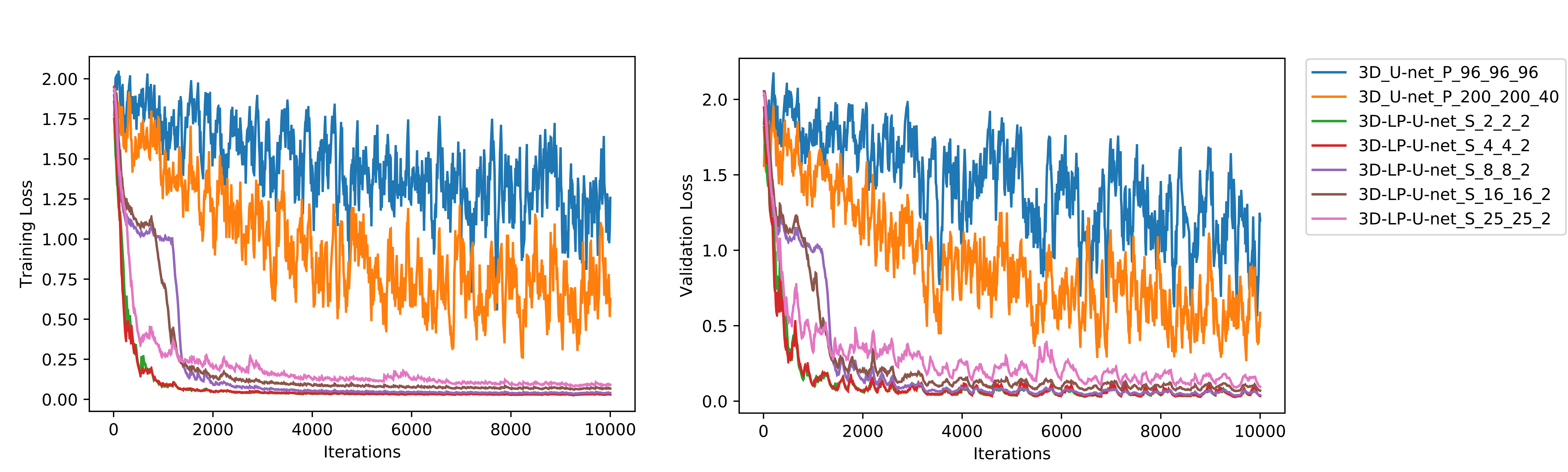}
\caption{Comparison of learning curves of the proposed 3D LP-U-Net with a fixed patch size of $400 \times 400 \times 80$ but different shuffling factors and the 3D U-net with different patch sizes. The left images shows the learning curves of the training data and the right image shows the learning curves of the validation data, where ``{\text{3D\_U-net\_P\_X\_Y\_Z}}'' means the results obtained from the 3D U-net with a patch size of $X \times Y \times Z$ and ``{\text{3D-LP-U-net\_S\_x\_y\_z}}'' means the results obtained from the 3D LP-U-net with a shuffling factor of (x, y, z). } 
\label{fig:LearningCurves}
\end{figure}

We further analyze the learning process of the proposed approach and the 3D U-net. As shown in Fig. \ref{fig:LearningCurves}, in all cases, as the training loss goes down, the validation loss decreases consistently, demonstrating that there is no serious over-fitting for all models even with such small datasets. From Fig. \ref{fig:LearningCurves}, we observe that due to the smaller patch size allowed by the 3D U-net, its learning curves are not smooth. Furthermore, the 3D U-net with large patch size has lower losses on both training and validation datasets than the one with small patch size, demonstrating the importance of using large patch size.  

\begin{figure}[tbp]
\centering
\includegraphics[width=\textwidth]{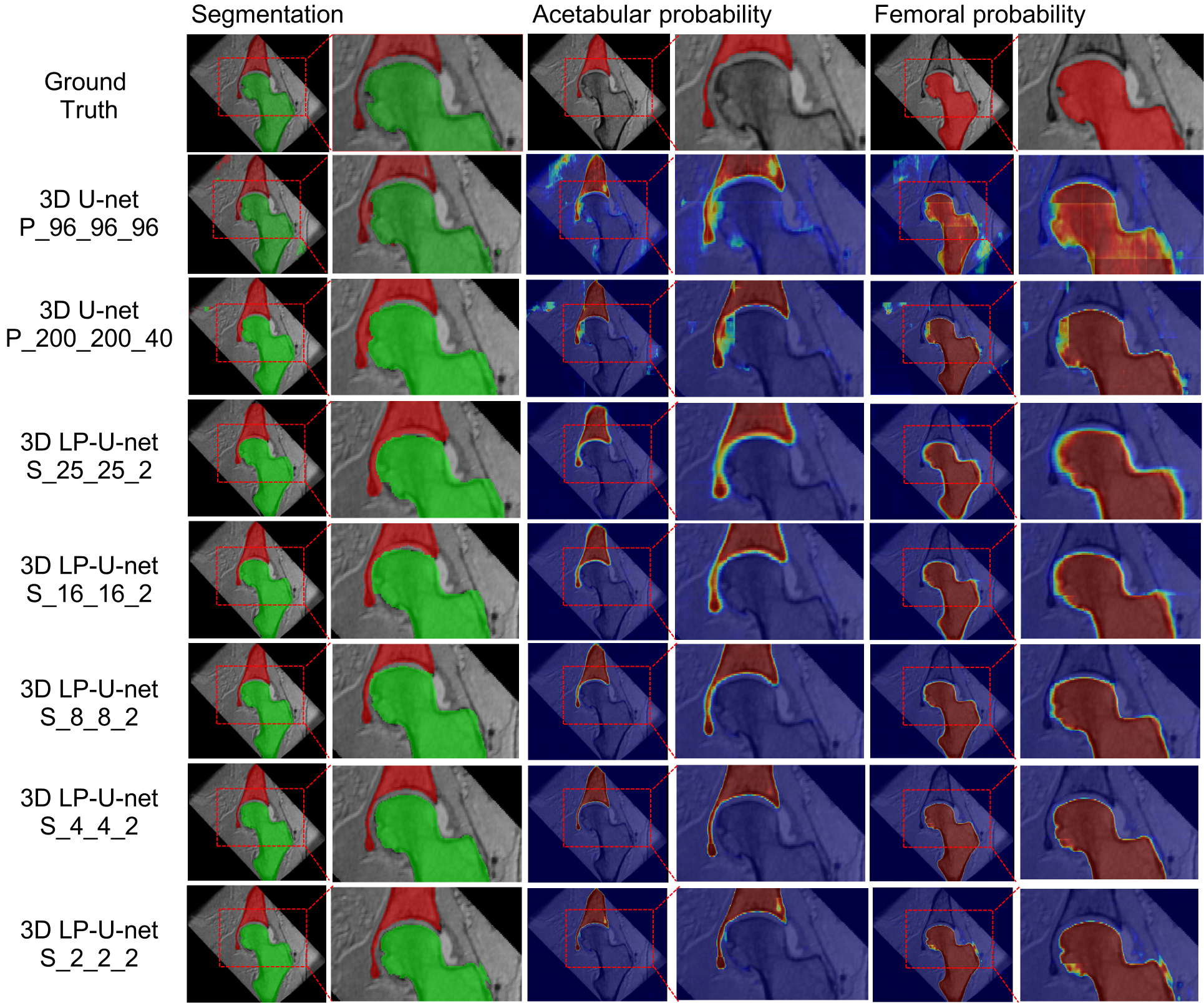}
\caption{Qualitative comparison of the segmentation results of 3D LP-U-Net with different shuffling factors and the 3D U-net with different patch sizes.}
\label{fig:QuanlitativeComparison}
\end{figure}

When comparing the learning curves of the 3D LP-U-net and the 3D U-net in Fig. \ref{fig:LearningCurves}, clear distinctions can be observed. First, due to the usage of large patch size, the learning curves of 3D LP-U-net are quite smooth. More importantly, the 3D LP-U-net not only converges much faster than the 3D U-net but also produces much lower losses on both training and validation datasets. It is also interesting to observe that for the 3D LP-U-net, in general, the bigger the shuffling factors, the larger the converged losses but the best results were obtained when the shuffling factor was (4, 4, 2). Such a qualitative observation was consistent with the quantitative results shown in Table \ref{Tab_3DLPU-net_ShufflingFactors}. These results also demonstrate that the proposed HDC can effectively speed up the training procedure by overcoming optimization difficulties via learning better context features from large patches.

Fig. \ref{fig:QuanlitativeComparison} visually compares the segmentation results obtained by the 3D LP-U-net with a fixed patch size of $400 \times 400 \times 80$ but different shuffling factors and the 3D U-net with different patch sizes. In this figure, we show both the overall segmentation and the probability of each structure as well as the results around the hip joint. From this figure, we observe that (1) less false positive segmentation was observed when comparing the results obtained by the 3D LP-U-net with those by the 3D U-net; and (2) for the 3D LP-U-net, the larger the shuffling factors, the higher the uncertainty around the boundary.

\begin{table}[tbp]
\centering
\caption{Results when different shuffling factors were used for the 3D LP-U-net. The size of the input patch is fixed to $400 \times 400 \times 80$.}
\label{Tab_3DLPU-net_ShufflingFactors}
\resizebox{\textwidth}{!}
{
\begin{tabular}{l|c|c|c|c|c|c|c|c|c|c} \hline
Shuffling factors & \multicolumn{2}{c|}{(2, 2, 2)}        & \multicolumn{2}{c|}{(4, 4, 2)}     & \multicolumn{2}{c|}{(8, 8, 2)}    & \multicolumn{2}{c|}{(16, 16, 2)}	& \multicolumn{2}{c}{(25, 25, 2)} \\ \hline
Initial learning rate & \multicolumn{2}{c|}{ 1.0E-03 } 						& \multicolumn{2}{c|}{ 2.0E-03 }			& \multicolumn{2}{c|}{ 3.0E-03 }				& \multicolumn{2}{c|}{ 5.0E-03 }		& \multicolumn{2}{c}{ 2.0E-02 } \\ \hline
Anatomy    & Ace               & Femur            & Ace              & Femur            & Ace              & Femur            & Ace              & Femur 					 & Ace              & Femur \\ \hline
DOC (\%)   & 96.77 $\pm$ 1.27  & 97.41 $\pm$ 1.34 & 96.77 $\pm$ 1.26 & 97.95 $\pm$ 0.63 & 96.30 $\pm$ 0.97 & 97.25 $\pm$ 0.59 & 94.24 $\pm$ 1.73 & 95.75 $\pm$ 1.02 & 91.57 $\pm$ 2.03 & 93.82 $\pm$ 1.52 \\ \hline
ASD (mm)   & 0.39 $\pm$ 0.28   & 0.43 $\pm$ 0.28  & 0.39 $\pm$ 0.28  & 0.33 $\pm$ 0.16  & 0.37 $\pm$ 0.11  & 0.41 $\pm$ 0.09  & 0.62 $\pm$ 0.23  & 0.64 $\pm$ 0.16  & 0.86 $\pm$ 0.25  & 0.96 $\pm$ 0.25 \\ \hline
HD (mm)    & 7.73 $\pm$ 3.81   & 6.23 $\pm$ 2.32  & 8.57 $\pm$ 5.68 & 3.59 $\pm$ 3.95   & 6.97 $\pm$ 3.15 & 5.15 $\pm$ 1.43   & 10.69 $\pm$ 7.44 & 6.39 $\pm$ 1.50  & 12.64 $\pm$ 2.87 & 8.18 $\pm$ 0.66 \\ \hline
\end{tabular}
}
\end{table}

\begin{table}[tbp]
\centering
\caption{Results when the original 3D V-net and the original HighRes3DNet were used with different patch sizes.}
\label{Tab_3DV-netHighRes3DNet}
\resizebox{\textwidth}{!}
{
\begin{tabular}{l|c|c|c|c|c|c|c|c}
\hline
\begin{tabular}[c]{@{}l@{}}Architectures \\ (Used patch size)\end{tabular} & \multicolumn{2}{c|}{\begin{tabular}[c]{@{}c@{}}3D V-net\\ $96 \times 96 \times 96$ \end{tabular}} & \multicolumn{2}{c|}{\begin{tabular}[c]{@{}c@{}}3D V-net\\ $200 \times 200 \times 40$ \end{tabular}} & \multicolumn{2}{c|}{\begin{tabular}[c]{@{}c@{}}HighRes3DNet\\ $100 \times 100 \times 80$ \end{tabular}} & \multicolumn{2}{c}{\begin{tabular}[c]{@{}c@{}}HighRes3DNet\\ $200 \times 200 \times 20$ \end{tabular}} \\ \hline
Anatomy                                                                    & Ace                                        & Femur                                      & Ace                                        & Femur                                     & Ace                                          & Femur                                       & Ace                                          & Femur                                       \\ \hline
DOC (\%)                                                                   & 88.71 $\pm$ 6.21                           & 92.27 $\pm$ 3.68                           & 92.78 $\pm$ 0.50                           & 96.67 $\pm$ 0.85                          & 90.66 $\pm$ 6.68                             & 86.18 $\pm$ 5.08                            & 93.04 $\pm$ 4.31                             & 93.58 $\pm$ 2.46                            \\ \hline
ASD (mm)                                                                   & 1.77 $\pm$ 1.29                            & 1.75 $\pm$ 0.77                            & 0.97 $\pm$ 0.97                            & 0.59 $\pm$ 0.23                           & 1.80 $\pm$ 2.36                              & 2.37 $\pm$ 0.59                             & 1.77 $\pm$ 2.34                              & 1.50 $\pm$ 0.82                             \\ \hline
HD (mm)                                                                    & 15.77 $\pm$ 6.16                            & 14.0 $\pm$ 3.70                           & 12.15 $\pm$ 6.81                           & 9.92 $\pm$ 4.26                           & 15.94 $\pm$ 11.70                            & 17.27 $\pm$ 4.93                            & 22.79 $\pm$ 13.92                            & 16.67 $\pm$ 6.53                            \\ \hline
\end{tabular}
}
\end{table}

\begin{table}[tbp]
\centering
\caption{Results when different shuffling factors were used for the 3D LP-V-net and the LP-HighRes3DNet.}
\label{Tab_3DLPV-netHighRes3DNet_ShufflingFactors}
\resizebox{\textwidth}{!}
{
\begin{tabular}{l|c|c|c|c|c|c|c|c|c|c}
\hline
\multicolumn{11}{c}{ Results of the 3D LP-V-Net with a fixed patch size of $400 \times 400 \times 80$ but different shuffling factors}                                                                                                                                                                              \\ \hline
Shuffling factors & \multicolumn{2}{c|}{(2, 2, 2)}      & \multicolumn{2}{c|}{(4, 4, 2)}      & \multicolumn{2}{c|}{(8, 8, 2)}      & \multicolumn{2}{c|}{(16, 16, 2)}    & \multicolumn{2}{c}{(25, 25, 2)}    \\ \hline
Anatomy           & Ace              & Femur            & Ace              & Femur            & Ace              & Femur            & Ace              & Femur            & Ace              & Femur            \\ \hline
DOC (\%)          & 95.58 $\pm$ 1.43 & 97.11 $\pm$ 0.63 & 94.98 $\pm$ 1.81 & 96.62 $\pm$ 0.38 & 93.21 $\pm$ 1.74 & 94.55 $\pm$ 0.88 & 91.66 $\pm$ 2.06 & 93.45 $\pm$ 1.40 & 90.05 $\pm$ 2.83 & 92.69 $\pm$ 1.47 \\ \hline
ASD (mm)          & 0.63 $\pm$ 0.58  & 0.49 $\pm$ 0.15  & 0.56 $\pm$ 0.34  & 0.51 $\pm$ 0.06  & 0.69 $\pm$ 0.25  & 0.82 $\pm$ 0.14  & 0.85 $\pm$ 0.27  & 1.0 $\pm$ 0.22   & 1.05 $\pm$ 0.36  & 1.10 $\pm$ 0.21  \\ \hline
HD (mm)           & 11.21 $\pm$ 9.97 & 7.24 $\pm$ 2.01  & 8.51 $\pm$ 4.33  & 5.97 $\pm$ 1.74  & 10.77 $\pm$ 6.88 & 6.76 $\pm$ 0.97  & 11.22 $\pm$ 4.76 & 7.85 $\pm$ 1.26  & 11.73 $\pm$ 6.62 & 7.48 $\pm$ 1.29  \\ \hline
\multicolumn{11}{c}{Results of the LP-HighRes3DNet with a fixed patch size of $400 \times 400 \times 80$ but  different shuffling factors}                                                                                                                                                                       \\ \hline
Shuffling factors & \multicolumn{2}{c|}{(4, 4, 1)}      & \multicolumn{2}{c|}{(4, 4, 2)}      & \multicolumn{2}{c|}{(8, 8, 2)}      & \multicolumn{2}{c|}{(16, 16, 2)}    & \multicolumn{2}{c}{(25, 25, 2)}    \\ \hline
Anatomy           & Ace              & Femur            & Ace              & Femur            & Ace              & Femur            & Ace              & Femur            & Ace              & Femur            \\ \hline
DOC (\%)          & 95.99 $\pm$ 1.18 & 97.38 $\pm$ 0.52 & 95.35 $\pm$ 1.30 & 96.62 $\pm$ 1.08 & 93.72 $\pm$ 1.69 & 95.52 $\pm$ 0.94 & 91.15 $\pm$ 2.09 & 92.41 $\pm$ 1.69 & 88.21 $\pm$ 2.48 & 90.0 $\pm$ 2.24  \\ \hline
ASD (mm)          & 0.43 $\pm$ 0.18  & 0.44 $\pm$ 0.12  & 0.53 $\pm$ 0.29  & 0.60 $\pm$ 0.33  & 0.66 $\pm$ 0.21  & 0.75 $\pm$ 0.19  & 0.90 $\pm$ 0.25  & 1.22 $\pm$ 0.32  & 1.27 $\pm$ 0.51  & 1.55 $\pm$ 0.36  \\ \hline
HD (mm)           & 8.21 $\pm$ 4.05 & 6.76 $\pm$ 2.68-  & 8.48 $\pm$ 4.33  & 7.85 $\pm$ 4.11  & 11.31 $\pm$ 4.47 & 8.38 $\pm$ 3.75  & 12.95 $\pm$ 7.14 & 9.53 $\pm$ 2.25  & 16.23 $\pm$ 5.12 & 9.62 $\pm$ 2.30  \\ \hline
\end{tabular}
}
\end{table}

Finally, we checked the influence of different architectures of the underlying FCNs on the performance of the proposed pipeline. Table \ref{Tab_3DV-netHighRes3DNet} shows the results when the original 3D V-net and the original HighRes3DNet were used with different patch sizes.  Please note that caused by high spatial resolution, HighRes3DNet \cite{Li_IPMI_2017} requires largest GPU memory to store intermediate results among all three architectures, though it has the smallest number of training parameters. Thus, the maximally allowed size of the input patch for the HighRes3DNet was $200 \times 200 \times 20$. In comparison, the results of the 3D LP-V-net and the LP-HighRes3DNet with different shuffling factors are reported in Table \ref{Tab_3DLPV-netHighRes3DNet_ShufflingFactors}. From the results reported in Tables \ref{Tab_3DLPU-net_ShufflingFactors}, \ref{Tab_3DV-netHighRes3DNet} and \ref{Tab_3DLPV-netHighRes3DNet_ShufflingFactors}, we can observe that (1) results achieved by the 3D LP-V-Net and the LP-HighRes3DNet are better than those achieved by the associated baseline when the chosen shuffling factor is not too big. For example, even with a shuffling factor of (8, 8, 2), the performance of the LP-HighRes3DNet is much better than that achieved by the original HighRes3DNet with the largest patch size allowed; and (2) the bigger the shuffling factor, the less accurate the results.

\begin{figure}[tbp]
\centering
\includegraphics[width=\textwidth]{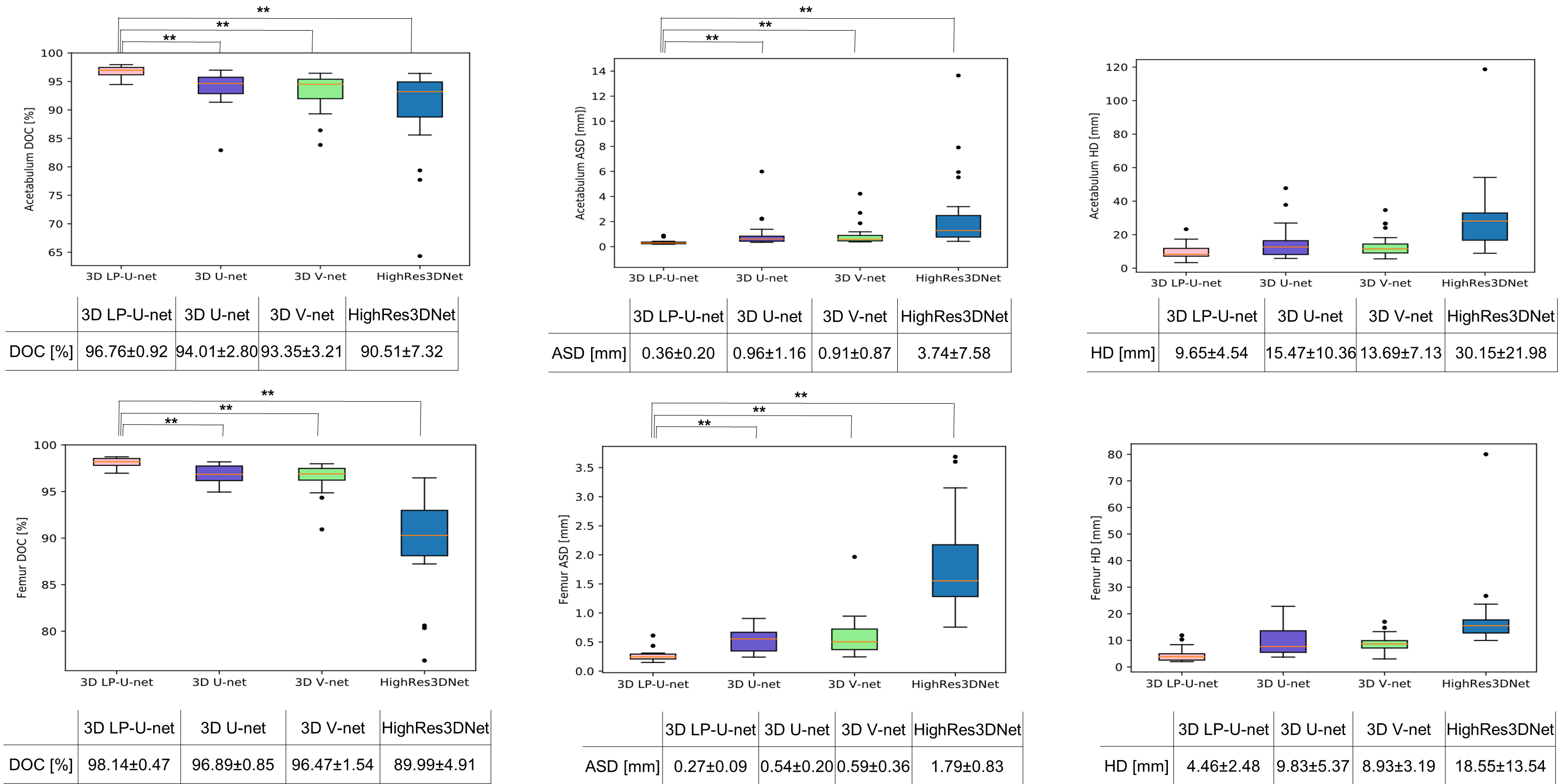}
\caption{Boxplots showing segmentation accuracy of the proposed 3D LP-U-net and three state-of-the-art methods. **indicates significant accuracy improvement with significance level of 0.01.}
\label{fig:HipMainStudyResults}
\end{figure} 

\subsection{Validation on hip MR images} 
We conducted a standard 5-fold cross validation study on the 25 T1 hip MR images with limited field of view. We used the same data augmentation strategy and the same training strategy as we used in the ablation study. In this study, for the 3D LP-U-net, we chose a fixed patch size of $400 \times 400 \times 80$ and a fixed shuffling factor of (4, 4, 2). We compared the  performance of the 3D LP-U-net with state-of-the-art methods such as 3D U-net \cite{3DUNET_2016}, 3D V-net \cite{Milletari_VNet_2016}, and HighRes3dNet \cite{Li_IPMI_2017}. For the 3D U-net and the 3D V-net, the chosen patch size is $200 \times 200 \times 40$ while for the HighRes3DNet, the patch size was chosen to be $200 \times 200 \times 20$.  The top row of Fig. \ref{fig:HipMainStudyResults} shows boxplots for overall DOC, ASD and HD of all four methods for segmenting the acetabulum. An average DOC of 96.76 $\pm$ 0.92\%, 94.01 $\pm$ 2.80\%, 93.35 $\pm$ 3.21\% and 90.51 $\pm$ 7.32\% was found for the 3D LP-U-net, the 3D U-net, the 3D V-net and the HighRes3DNet, respectively. The 3D LP-U-net showed significantly higher accuracy than all other three methods ($p < 0.01$). For ASD, the same significance was also observed. The bottom row of Fig. \ref{fig:HipMainStudyResults} shows the comparison results for the proximal femur. An average DOC of 98.14 $\pm$ 0.47\%, 96.89 $\pm$ 0.85\%, 96.47 $\pm$ 1.54\% and 89.99 $\pm$ 4.91\% was found for the 3D LP-U-net, the 3D U-net, the 3D V-net and the HighRes3DNet, respectively. The 3D LP-U-net showed significantly higher accuracy than all other three methods ($p < 0.01$) when segmenting the proximal femur. These results demonstrate the efficacy of the proposed approach.        

\subsection{Validation on MICCAI 2015 IVD localization and segmentation challenge data }
We conducted experiments on the MICCAI 2015 IVD localization and segmentation challenge data \cite{Zheng_MedIA_2017}, which contains 25 3D T2-weighted MR images. The resolution of all images were resampled to $2mm \times 1.25mm \times 1.25mm$. The size of the images is between $39 \times 305 \times 305$ and $48 \times 304 \times 304$ voxels. Each image contains at least 7 IVDs T11-S1. These 25 MR images were divided into three non-overlapped subsets as training data (15 3D MR images), Test1 data (5 3D MR images) and Test2 data (the remaining 5 3D MR images). All methods were trained on the training data and then separately evaluated on the two testing datasets. Manual segmentation was used as the reference for all evaluations.

We compared the performance of the 3D LP-U-net with top-5 state-of-the-art methods described in \cite{Zheng_MedIA_2017}. In the training phase, we chose a fixed patch size of $32 \times 288 \times 288$ voxels and a fixed shuffling factor of (1, 2, 2) for the 3D LP-U-net in order to incorporate as large as possible context information. Table \ref{Tab_IVDSegResults} shows the accuracy comparison between 3D LP-U-net and the state-of-the-art methods as described in \cite{Zheng_MedIA_2017}. For both testing datasets, the 3D LP-U-net achieved consistently better results than other state-of-the-art methods. It is worth to mention that the 3D LP-U-net outperforms the method from the team UNICHK, which is a deeply supervised 3D segmentation network, by nearly 3.6\% in terms of average DOC, which is a large improvement. The lower standard deviation of DOC shows that the 3D LP-U-net is the most stable and robust across all different IVD cases. The results that we obtained proves the effectiveness of our approach.

\begin{table}[tbp]
\caption{Accuracy (DOC, \%) comparison between the 3D LP-U-net and the state-of-the-art methods as described in \cite{Zheng_MedIA_2017}. }
\label{Tab_IVDSegResults}
\begin{tabular}{l|c|c|c}
\hline
Method      & \multicolumn{1}{l|}{Test1 results (\%)} & \multicolumn{1}{l|}{Test2 results (\%)} & \multicolumn{1}{l}{Average (\%)} \\ \hline
3D LP-U-net & \textbf{92.4 $\pm$ 1.5}                         & \textbf{92.1 $\pm$ 1.7}                                   & \textbf{92.2 $\pm$ 1.7}                   \\ \hline
UNILJU      & 91.5 $\pm$ 2.3                         & 92.0 $\pm$ 1.9                                   & 91.8 $\pm$ 2.1                   \\ \hline
UNIBE       & 89.8 $\pm$ 2.9                         & 91.2 $\pm$ 2.0                                   & 90.5 $\pm$ 2.6                   \\ \hline
UNIEXE      & 89.8 $\pm$ 3.6                         & 90.2 $\pm$ 2.6                                   & 90.0 $\pm$ 3.1                   \\ \hline
Sectra      & 90.0 $\pm$ 2.6                         & 90.0 $\pm$ 2.2                                   & 90.0 $\pm$ 2.4                   \\ \hline
UNICHK      & 88.4 $\pm$ 3.7                         & 88.9 $\pm$ 3.4                                   & 88.6 $\pm$ 3.5                   \\ \hline
\end{tabular}
\end{table}

\section{Conclusion}
We proposed a simple yet effective holistic decomposition convolution for improving semantic segmentation systems. The HDC consists of a periodic down-shuffling operation followed by a conventional 3D convolution. It can be directly applied to the input data and has the advantage of significantly reducing the size of the data for sub-sequential processing while using all the information available in the input irrespective of the down-shuffling factors. To achieve volumetric dense prediction at the output, we used a previously introduced dense upsampling convolution. We showed that HDC and DUC were network agnostic and could be combined with different FCNs for an improved performance. Experimental results demonstrated the effectiveness of our framework on different semantic segmentation tasks. 

%
%
\label{sect:bib}

\end{document}